# Applying Ensemble Methods to Model-Agnostic Machine-Generated Text Detection

Ivan J.E. Ong
*Georgia Institute of Technology*
*Atlanta, GA, United States*
iong7@gatech.edu

Boon King Quek
*Georgia Institute of Technology*
*Atlanta, GA, United States*
bquek3@gatech.edu

**Abstract**

*In this paper, we study the problem of detecting machine-generated text when the large language model (LLM) it is possibly derived from is unknown. We do so by apply ensembling methods to the outputs from DetectGPT classifiers (Mitchell et al. 2023), a zero-shot model for machine-generated text detection which is highly accurate when the generative (or base) language model is the same as the discriminative (or scoring) language model. We find that simple summary statistics of DetectGPT sub-model outputs yield an AUROC of 0.73 (relative to 0.61) while retaining its zero-shot nature, and that supervised learning methods sharply boost the accuracy to an AUROC of 0.94 but require a training dataset.* This *suggests the possibility of further generalisation to create a highly-accurate, model-agnostic machine-generated text detector.*

## 1. Introduction

In this paper, we extend the DetectGPT model developed in Mitchell et al. (2023), by relaxing the constraint that the generative (or base) model and the discriminative (or scoring) model are the same, mirroring real-life scenarios where the base model is unknown. The DetectGPT paper briefly explores this issue and finds that detection performance is expectedly lower and with significant variations in accuracy depending on the base and scoring model. We take an ensemble approach and combine the outputs of multiple DetectGPT classifiers, each assuming a different base model, using supervised learning on these features to create effective classifiers that are agnostic to the LLM that the text is potentially generated from.

Current approaches to the machine-generated text detection problem make assumptions that the text was generated by a particular base model, to which the discriminator has different levels of access depending on the problem setting. We adopt the classification approach outlined by OpenAI (Solaiman et al., 2019).

The *black-box setting* presumes only access to the model outputs. Black-box approaches mainly consist of creating stylometric features such as TF-IDF (term frequency-inverse document frequency), skip-gram, continuous-bag-of-words (Reddy et al., 2020) and fitting one or numerous binary classifiers to them. These can range from logistic regression models to convolutional neural networks (Weller and Woo, 2019) or LSTM models (Kudugunta and Ferrara, 2018). These binary classifiers can also act as base learners in ensemble methods (Fayaz et al., 2020). These features can also be augmented with additional information such as account data in the context of social media bot detection. However, high classification accuracy for these methods are reliant on sufficiently-long text length and a sufficiently-diverse corpus of training machine-generated samples in terms of stylometric and linguistic characteristics in order to prevent overfitting. As such, these classifiers need to be continually trained and updated, limiting their usefulness (Pegoraro et al., 2023).

In contrast, the *white-box* setting repurposes the generative model as a classifier, and provides the discriminator access to some or all of the model's inner attributes, such as log-probabilities and scores. White-box approaches can be further classified into two subcategories: i) zero-shot models which are able to discriminate without explicit training data, and ii) fine-tuning based models, which require domain-specific labelled training texts to work. Under the assumption that the discriminative model is also the generative model, white-box approaches tend to be highly effective even using simple metrics such as the average log probability of all tokens within the candidate text (Solaiman et al., 2019). White-box models include GPTZero, which uses perplexity based on GPT-2 (Gold Penguin, 2023) and burstiness as key metrics.

The increasingly articulate and realistic nature of texts generated by LLMs have enabled their use in applications such as drafting essays, summarising information, as well as writing code proof-of-concepts. This also raises concerns about factual inaccuracies in journalism (The Verge, 2023), the potential corruption of technical knowledge (Rodriguez et al., 2022), misattribution of work extending to academic plagiarism, and the facilitating of phishing techniques and other scams. Researchers have hence recently been tackling the detection problem of identifying whether a candidate piece of text is machine or human-generated. However, this is complicated by the rapidly-evolving LLM landscape, with new models being frequently rolled out and the development of adversarial methods to circumvent detection methods. Our paper hence makes a contribution by showing how simple ensembling methods can easily improve the effectiveness of machine-generated text detection without assuming the base model. This is of practical use in real-world situations such as fake news and spam detection, where a threat actor uses a single, unknown base model to propagate deceptive information at scale.

We use two datasets to evaluate the effectiveness of our methods. The first is the Extreme Summarization (XSum) dataset (Narayan et al., 2018), which consists of English news articles extracted from the British Broadcasting Corporation (BBC) spanning all topics from 2010 to 2017 to evaluate the cross-domain effectiveness of our methods. The second is the German translations of news articles from the WMT16 dataset (Bojar et al., 2016) to analyse the transferability of our methods to languages other than the one the scoring models have been trained on (i.e. English).

## 2. Approach

### 2.1 Summary of DetectGPT

As our work is an extension of DetectGPT, developed by Mitchell et al. (2023), we provide an overview of it works. DetectGPT is distinct from other white-box detectors in that it estimates log-probabilities of semantically similar texts instead of just the sample text. The key observation is the hypothesis that LLMs implicitly choose samples to maximise the log-probability of successive tokens when generating text, and hence tend to sample from regions of negative curvature of the log-probability function of p_theta. Hence, instead of using just raw log-probabilities of the original text as done in previous work, one can instead perturb an original text $x$ into a rewritten text $x'$ which is semantically similar to $x$, using an off-the-shelf mask-filling model such as T5 (Raffel et al., 2019). Taking the difference in log-probabilities between the original and perturbed text $log(\frac{p_\theta(x)}{p_\theta(x')})$ and averaging over multiple perturbed texts $x'_n$ yields the perturbation discrepancy $d$, which is an estimate of the local curvature of the sample's log-probability function in some latent semantic space. Mitchell et al. (2023) empirically show that $d$ is positive and **larger** for text generated by the base model, and thus an effective way of discriminating between human-written and machine-generated text. They also find that the normalised perturbation discrepancy $z = \frac{d - \mu_d}{\sigma_d}$ yields slightly higher accuracy in terms of the Area under the Receiver Operating Curve (AUROC).

### 2.2 Motivating the Use of Ensembling Methods

We use ensembling methods to generalise DetectGPT's detection capabilities assuming a *particular* base model to *any* model. Our methodology is summarised in *Figure 1*. The hypothesis behind ensembling is that DetectGPT under a base model A acts as a surrogate to provide a weak signal as to whether a sample text is generated from *any other* model B; the more closely related their architectures, the stronger the signal. In general, the more diverse the classifier ensemble, the better it performs (Kuncheva, L. and Whittaker, C., 2003). There are already similar ensemble methods in the black-box domain, with researchers having applied a wide range of classical text metrics such as TF-IDF in predictive models such as neural networks and aggregating them to produce accurate classifiers (Reddy et al., 2020). We hence thought that applying well-established methods to DetectGPT, a relatively recent detection model, in order to relax its constraints, would be a novel contribution and provide meaningful results (whether positive or negative).

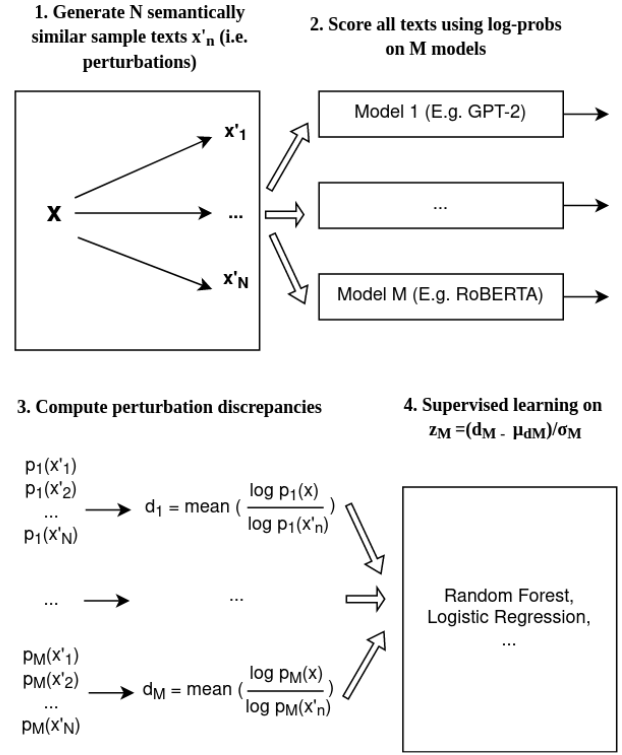

*Figure 1*. We modify the DetectGPT model used in Mitchell et al. by perturbing N texts (Step 1), running each of the original and perturbed texts through M DetectGPT sub-models (Step 2), each with a distinct LLM, and using the M perturbation discrepancies $d_m$ in simple summary statistics and supervised-learning models (Steps 3 and 4).

We encountered some initial difficulties when fine-tuning the research scope. Since we wanted to improve on detection techniques, we had to search for a recent model with open-source implementation code. One of our first options was refining GPTZero, which uses conventional perplexity and burstiness metrics to classify text, but it is unfortunately closed-source and we found it difficult to obtain reliable code. After doing more research, we found DetectGPT, which we found to have promising directions for further research and which had a well-maintained code base. Another more conceptual hurdle was that the supervised learning models we planned to use produce binary classifications; hence, we had to find a way to output a continuous probability or confidence value to be consistent with the AUROC metric. We found that the predict_proba() function in scikit-learn furnished us with the class probabilities. Another practical limitation was our limited compute power relative to the original authors of the DetectGPT paper. We could only reliably utilise mask-filling and scoring models with at most 400 million parameters, limiting the chance of our methods to achieve high accuracy. However, we felt that it was still worthwhile to use smaller models, as they could still serve as 'lower bounds' on the potential classification accuracy.

We modify the official DetectGPT implementation provided by Mitchell et al. (2023). To facilitate our experiments, we optimised the existing code by reusing the same samples and perturbations for all combinations of models, resulting in a 5x speedup in runtime, as well as writing a shell script to iterate through all combinations to be analysed and a utilities library to abstract out the common logic of results extraction and aggregation for rapid analysis.

### 2.3 Categorisation of Methods

Our methods of combining outputs fall into three categories: 1) Baseline results from the individual sub-models, which are self-explanatory; 2) Ensembling using simple summary statistics; 3) supervised learning models.

Category 2 consists of using the arithmetic mean, median and maximum of the scores of all the sub-models. Using the mean mimics soft-voting and using the median mimics hard-voting. The intuition behind taking the maximum is that each individual classifier is assessing the likelihood that the sample text is generated by their model; as long as one sub-model shows a relatively high z-score, the sample text is likely to be machine-generated.

Category 3 consists of supervised learning methods. We consider the following models: logistic regression model, random forest, a Naive-Bayes Gaussian model and support vector machine. While these methods might have higher accuracy, we also implicitly make a tradeoff by discarding the zero-shot aspect of DetectGPT due to the requirement for a training dataset for our aggregating model. This class of methods thus assumes that all machine-generated samples are derived from the same (possibly unknown) model.

We also attempt a "multi-stage" model that orders the DetectGPT sub-models by decreasing complexity and successively considers each sub-model, which we classify under Category 3. If the *d*-score is below a threshold, which we set as $\mu_D - z_{opt}\sigma_D$, where $\mu_D$ and $\sigma_D$ are taken over the training set, and $z_{opt}$ is adaptively set for each sub-model to maximise accuracy on the training set, we halt and take the mean of all sub-models considered so far. This approach aims to reduce the likelihood of false positives by combining the idea of taking the maximum of all DetectGPT models with the hypothesis that more complex scoring models, by virtue of having more representational power to "encompass" sub-models, are more accurate at evaluating if a text is machine-generated.

## 3. Experiments and Results

### 3.1 Experiments

For each dataset, we generate synthetic samples from the raw conditional distribution of the corresponding base model using the first 30 tokens as input, with a temperature of 1 to ensure reproducibility of samples. In general, we use smaller models than current state-of-the-art LLMs, as they are likely to require less compute time and thus not aggravate the long compute time, which is one of the noted limitations of DetectGPT method. We choose to focus on causal language models for the base models, which are typically used for text generation. We use T5-base as our mask-filling model due to compute limitations, and because the rewritten samples produced by it are sufficiently realistic. Our scoring models comprise of our base models and two additional LLMs to provide more features. The finalised set of base, mask-filling, scoring models used in our experiments is shown in *Table 1*, although our methodology can be easily extended to encompass additional models. Because our focus is on ensembling methods, we reuse the same set of hyperparameters for all experiments as recommended by Mitchell et al. (2023), such as masked span length (2), fraction of words masked for filling (0.15) and number of perturbations (50).

| Base | Mask-Filling | Scoring |
|---|---|---|
| gpt-neo (125m) gpt2 (124m) gpt2-medium (355m) | T5-base (220m) | gpt-neo gpt2 gpt2-medium bert-base-cased (110m) RoBERTA-base (125m) |

*Table 1.* The combinations of DetectGPT sub-models used in our ensembling methods. The number of parameters (in millions) are shown in brackets for each model when it first appears.

We perform hyperparameter tuning using grid search for all four models, which are implemented in Scitkit-learn. We find in practice that the default settings work well without requiring further tuning, with marginal improvements in accuracy. *Table 2* shows the set of hyperparameters tuned for each supervised learning method.

We use AUROC as our accuracy metric, which can be interpreted as the probability that a random human-written text is ranked lower (i.e. is more "human") than a random machine-generated text and which incorporates both the false positive and false negative rate.

| Logistic Regression | Random Forest | Gaussian Naive Bayes | Support Vector Machine |
|---|---|---|---|
| C penalty | bootstrap max_depth min_samples_split n_estimators | var_smoothing | C gamma kernel |

Table 2. Combinations of hyperparameter tuning for the supervised learning methods (Category 3).

## 3.2 Results

Our results for the three categories are shown in *Tables 3, 4* and *5* respectively. Accuracies of above 0.9 are shown in green for all tables. *Table 3* confirms the hypothesis that DetectGPT's accuracy works best when the scoring model is also the base model. Interestingly, gpt2-medium, despite having the largest number of parameters, is not the most accurate, although it scores the best at detecting instances generated from itself; instead, gpt2 is the most accurate.

| Base Model | gpt-neo | gpt2 | gpt2-medium | BERT | RoBERTa |
|---|---|---|---|---|---|
| gpt-neo | **0.95** | 0.82 | 0.78 | 0.37 | 0.47 |
|  | **1** | 0.7 | 0.25 | 0.63 | 0.84 |
| gpt2 | 0.28 | **0.99** | 0.84 | 0.27 | 0.56 |
|  | **0.93** | **0.98** | 0.54 | 0.56 | 0.76 |
| gpt2-medium | 0.31 | **0.97** | **0.98** | 0.28 | 0.59 |
|  | **0.9** | **0.9** | **0.93** | 0.55 | 0.65 |
| Average | 0.73 | 0.89 | 0.72 | 0.45 | 0.64 |

Table 3. Baseline results (Category 1), with rows corresponding to the base model and columns corresponding to the scoring model. The first and second rows for each base model correspond to the German and XSum datasets respectively. As expected, DetectGPT is not model-agnostic.

*Table 4* shows results of using simple summary statistics. The experiments here are those where the base model is *excluded* from the set of scoring models. It shows that simple summary statistics taken across 'incorrect' sub-models (in that the scoring model is not the base model) still perform better than using just one 'incorrect' sub-model. This implies the potential of other simple estimators to obtain higher accuracy without the need for training.

| Base Model | Dataset | Baseline | Max | Mean | Median |
|---|---|---|---|---|---|
| gpt-neo | German | 0.61 | 0.63 | 0.73 | 0.77 |
|  | XSum | 0.60 | 0.32 | 0.72 | 0.74 |
| gpt2 | German | 0.49 | 0.55 | 0.43 | 0.45 |
|  | XSum | 0.70 | 0.56 | 0.88 | **0.91** |
| gpt2-medium | German | 0.54 | 0.78 | 0.68 | 0.64 |
|  | XSum | 0.75 | **0.9** | **0.93** | 0.88 |
| Average |  | 0.61 | 0.62 | 0.73 | 0.73 |

Table 4. Results for simple summary statistics (Category 2) when the base model is excluded from the set of scoring models. We compare these results, with the 'Baseline' column, which is the average accuracy of the DetectGPT sub-models, not including the base model.

Although our accuracies are lower than those in Mitchell et al. (2023), the model presented here is zero-shot and similarly agnostic to the base model, broadening its applicability. We note that the maximum performs the worst out of the three statistics, as it only uses information from the sub-model that is most confident about the synthetic nature of the sample text and ignores the rest, while the other two incorporate information from all sub-models. Furthermore, as these statistics can be computed without any training, these classifiers retain the zero-shot nature of DetectGPT while improving overall model-agnostic accuracy).

*Table 5* shows the results for Category 3, supervised learning methods. These methods generally perform the best, with models like logistic regression and support vector machines attaining an average accuracy of about 0.94. even with the base model being excluded. Including the experiments where the base model is one of the scoring models increases the accuracy further to 0.96. We compare our results to those we would have gotten from the unmodified DetectGPT setting, where the base model is the source model. Although our models do slightly worse than DetectGPT overall, the latter imposes a stricter assumption and thus this is to be expected. In fact, for (GPT2-Medium, XSum), the accuracy of GPT2-medium is *lower* than that of models that did not include it as a sub-model. However, we note that our multi-step estimator performs significantly worse than the other supervised methods, with it performing particularly poorly on the German dataset.

| Base Model | Dataset | DetectGPT | LR | RF | NB | SV | MS |
|---|---|---|---|---|---|---|---|
| gpt-neo | German | **0.95** | 0.78 | 0.80 | 0.77 | 0.82 | 0.74 |
|  | XSum | **1** | **0.98** | **0.97** | **0.95** | **0.98** | 0.36 |
| gpt2 | German | **0.99** | **0.93** | **0.92** | **0.90** | **0.93** | 0.84 |
|  | XSum | **0.98** | **1.0** | **0.99** | **0.98** | **1.0** | 0.67 |
| gpt2-medium | German | **0.98** | **0.99** | **0.98** | **0.98** | **0.98** | 0.38 |
|  | XSum | **0.93** | **0.96** | **0.95** | **0.93** | **0.96** | **0.93** |
| Average |  | 0.97 | 0.94 | 0.93 | 0.92 | 0.94 | 0.65 |

Table 5. Results for supervised learning methods (Category 3). We compare these results with the DetectGPT column, which shows the accuracy in the baseline DetectGPT setting, to which it is most suited. For the other columns showcasing our methods, the base model is excluded from the set of scoring models.

We also note that the coefficient for the fitted logistic regression for all experiments is positive and large for GPT-2, implying that the model is a relatively good discriminator for a large range of LLMs. This is consistent with our results in Table 3, where GPT2 was the best standalone sub-model.

## 3.3 Limitations

In addition to the known limitations of DetectGPT, which are its relatively long compute time and need for access to the log-probabilities of the scoring model, our current ensembling methods are performed on relatively few datasets and few models; it is likely that the greater the number and the more diverse the variety of DetectGPT sub-models used, the broader the model applicability. As such, more work can be done to validate the approach on other datasets, as well as to create a 'minimal covering set' of LLMs that when combined with our method can reliably and accurately classify any sample text. Model diversity will likely be a key consideration in creating this overarching set. Furthermore, while simply taking the mean and maximum (Category 2) can yield slight increases in accuracy relative to the individual DetectGPT sub-models, supervised learning methods (Category 3) require labelled training data to obtain further accuracy improvements. This imposes the assumption that machine-generated text within the training data must be derived from the same base model (which can be unknown). Future work can study the interplay between the number and diversity of sub-models and the volume of training data required to attain a required level of accuracy.

# Appendix 1: Division of Work

| Student Name | Contributed Aspects | Details |
|---|---|---|
| Ivan | Literature Review, Implementation | <ul><li>Perform literature review of existing detection techniques</li><li>Maintain and run experiment runner script permute.sh to permute over different combinations of model and experimental parameters, and generate experiment results for analysis</li><li>Improve, optimize and test main experiment program run.py</li><li>Maintain library of common helper functions (e.g. for retrieval, collation of results) in utils.py to aid analysis</li><li>Write code to perform supervised learning methods (logistics regression, random forest) and simple aggregation methods (mean, median, max) to ensemble the curvature output of different scoring models (see supervised.ipynb)</li></ul> |
| Boon King | Literature Review, Coding, Analysis | <ul><li>Perform literature review of existing detection techniques and propose extending DetectGPT paper</li><li>Write code to perform additional supervised learning methods (Naive Bayes, Support Vector Machine)</li><li>Propose and implement multi-step estimator</li><li>Perform hyperparameter tuning for supervised learning methods</li><li>Analyse and present significant findings for each category of methods</li></ul> |